\documentclass[journal]{IEEEtran}
\usepackage{hyperref}
\usepackage{color}
\usepackage{tikz}
\usepackage{colortbl}
\usepackage{xcolor}
\usepackage{multirow}
\usepackage{booktabs}

\usepackage{graphicx}
\usepackage{graphics,graphicx}

\graphicspath{{figures/}}

\usepackage{rotating}
\usepackage{float}
\usepackage{lineno}
\usepackage{epsfig}
\usepackage[linesnumbered,boxed,ruled]{algorithm2e}
\usepackage{amsmath}
\usepackage{amssymb}
\usepackage{multirow}
\usepackage{cases}
\usepackage{array}
\usepackage{url}
\usepackage[tight,footnotesize,sf,SF]{subfigure}
\usepackage{color}

\newcommand{\fref}[1]{Figure \ref{#1}}
\newcommand{\sref}[1]{Section \ref{#1}}
\newcommand{\tref}[1]{Table \ref{#1}}
\newcommand{\eref}[1]{Eq. (\ref{#1})}

\begin{document}

\title{Deep Reinforcement Learning for Online Routing of Unmanned Aerial Vehicles with Wireless Power Transfer}

\author{Kaiwen~Li,
        Tao~Zhang,
       Rui~Wang*~\IEEEmembership{IEEE Member},
       and Ling~Wang.
\thanks{This work was supported by the National Science Fund for Outstanding Young Scholars (62122093), the National Natural Science Foundation of China (72071205) and the Ji-Hua Laboratory Scienctific Project (X210101UZ210).}
\thanks{Kaiwen Li, Tao~Zhang, and Rui~Wang (corresponding author), are with the College of System Engineering, National University of Defense Technology, Changsha 410073, PR China (e-mail: kaiwenli\_nudt@foxmail.com, ruiwangnudt@gmail.com, zhangtao@nudt.edu.cn).
}
\thanks{Ling Wang is with Department of Automation, Tsinghua University, Beijing, 100084, P.R. China. }

}

\markboth{Journal of \LaTeX\ Class Files,~Vol.~14, No.~8, April~2022}%
{Deep Reinforcement Learning for Online Routing of Unmanned Aerial Vehicles with Wireless Power Transfer}
\maketitle

\begin{abstract}
The unmanned aerial vehicle (UAV) plays an vital role in various applications such as delivery, military mission, disaster rescue, communication, etc., due to its flexibility and versatility. This paper proposes a deep reinforcement learning method to solve the UAV online routing problem with wireless power transfer, which can charge the UAV remotely without wires, thus extending the capability of the battery-limited UAV. Our study considers the power consumption of the UAV and the wireless charging process. Unlike the previous works, we solve the problem by a designed deep neural network. The model is trained using a deep reinforcement learning method offline, and is used to optimize the UAV routing problem online. On small and large scale instances, the proposed model runs from four times to 500 times faster than Google OR-tools, the state-of-the-art combinatorial optimization solver, with identical solution quality. It also outperforms different types of heuristic and local search methods in terms of both run-time and optimality. In addition, once the model is trained, it can scale to new generated problem instances with arbitrary topology that are not seen during training. The proposed method is practically applicable when the problem scale is large and the response time is crucial.

\end{abstract}

\begin{IEEEkeywords}
Unmanned Aerial Vehicle, Deep Neural Network, Routing Optimization, Deep Reinforcement Learning.
\end{IEEEkeywords}

\IEEEpeerreviewmaketitle

\section{Introduction}

Nowadays, unmanned aerial vehicle (UAV) plays an important role in various fields, and it will get more usage in the future. Due to its flexibility and versatility, UAV has found a large number of applications, such as package delivery for civilian purposes \cite{wang2019vehicle}, search and rescue in natural disasters \cite{de2014uas}, surveillance and attack for military purposes \cite{bry2012state}, on-site inspection for industrial facilities \cite{nikolic2013uav}, providing mobile relay for the wireless communication \cite{chen2017optimum}, etc. 

One of the main limitations of UAV is its limited battery capacity \cite{simic2015investigation}. A  UAV in general can only work for dozens of minutes, which to some extent forbids the complete unmanned operation of the UAV. UAVs have to land to get battery charged when they run out of electricity. In addition, when it is not safe or convenient for the UAV to land, such as disaster rescue and military usage, the restricted working time of the UAV greatly limits its capability. 

Wireless power transfer (WPT), originally investigated by Nikola Tesla \cite{simic2015investigation} in the early twentieth century, can transmit the energy from the power supply to the target via the air instead of wires. With the development of WPT, the UAV can be powered remotely. Recently in December 2021, the American Defense Advanced Research Projects Agency (DARPA) has funded Electric Sky company to investigate the WPT technology to power up the UAV. This promising technology would significantly expand the capability of the UAV in plenty of applications, such as search, rescue, and surveillance in extreme environment and complete unmanned operation. 

In this context, we investigate the UAV routing problem with wireless power transfer technology. The UAV routing problem aims to find a flying route with minimum traveling time to visit a set of locations to complete a task, and go back to the base station once the UAV runs out of electricity. The base station starts to charge the UAV when they are close to each other. The UAV hovers at the top of the base station until it is fully charged. The whole process is completely unmanned, thus significantly reducing the labor cost and safety risks in extreme environment.  

The UAV routing problem is combinatorial in nature, which results in NP-hardness. Existing approaches for solving the routing problem mainly include the following three categories:

\begin{enumerate}
\item Exact algorithms, such as branch \& bound and dynamic programming, aim to search the whole solution space and provide the optimal solution for users. However, they suffer from heavy computational burden, and are usually intractable for large scale problems. 
\item Approximate algorithms can obtain near-optimal solutions with theoretical guarantees of the solution quality. But such algorithms do not exist for all problems, and the run-time can be still intractable when the time complexity is higher-order polynomial.
\item Heuristic algorithms, such as local search and evolutionary algorithms, run much faster than exact and approximate solvers. However, heuristic algorithms have no theoretical guarantee of the quality of their obtained solutions. Moreover, specialized domain knowledge and trial-and-error experiments are required to design such algorithms. As search-based algorithms, heuristic algorithms also suffer significantly long run-time for large scale problems where the solution space is too large for the algorithms to search. 
\end{enumerate}

Large scale problems arise frequently in practice, and system response time is especially crucial in various applications, such as search and rescue in disasters. However, it is hard for traditional combinatorial optimization solvers to achieve fast system response. Their computation time may increase exponentially with the problem dimension. New online optimization approaches are required to solve the UAV routing problems in practice. 

Recent advances in deep learning make the online optimization of large scale routing problems possible. Google Brain first proposed a deep neural network model \cite{bello2016neural} to solve combinatorial optimization problems in 2016. The model is trained offline while used online, thus reducing the online execution time significantly. Several approaches follow this line of work, and some of them are shown as  effective in solving various routing problems. Traditional approaches design heuristics for  combinatorial optimization by human. Instead, deep learning method learns heuristics from data, i.e., a collection of instances from a certain type of problem. It provides a new way to solve combinatorial optimization problems. 

In this study, we propose a deep reinforcement learning method to solve the UAV online routing problem with WPT. A deep neural network model with Encoder-Decoder structure is designed to map from the problem inputs to the UAV routes. Deep reinforcement learning method is used to optimize the neural network parameters. The contributions are as follows:

\begin{enumerate}
\item We investigate the UAV online routing problem with wireless power transfer using the deep reinforcement learning method. In comparison with traditional solvers that rely on handcrafted expert heuristics, the proposed method can exploit statistical similarities and learn data-driven heuristics from a collection of problem instances. 
\item Once the model is trained offline, it can solve different problem instances with arbitrary topology that are not seen in the training set. 
\item The proposed approach outperforms the state-of-the-art Google OR-Tools solver for combinatorial optimization. It runs from more than four times to 500 times faster than Google OR-Tools with nearly identical solution quality.
\item The proposed approach outperforms other heuristic and local search algorithms in terms of both run-time and optimality on all of the test instances.

\end{enumerate}

The remainder of the paper is organized as follows. We outline the related works in \sref{related}. The system model is introduced in \sref{system}, including the UAV model and the wireless power transfer model. The proposed deep neural network model is presented in \sref{smodel}, and the reinforcement learning method for training the model is introduced in \sref{train}. \sref{exp} presents the experimental settings and numerical results of the experiments. Concluding remarks and future perspectives are given in the last section. 

\section{Related Works}\label{related}

The UAV routing problem can be considered as a generalized version of traveling salesman problem (TSP) and vehicle routing problem (VRP). Concorde, LKH3, Google OR-Tools are well-known solvers that can effectively solve the routing problems like TSP and VRP. Note that Concorde and LKH3 are less flexible than Google-OR-Tools, since their input format is fixed and it is hard to add complex constraints into the problem. Google OR-Tools can solve a variety of routing problems and can support customized problem settings. Moreover, a number of heuristic approaches that consider additional constraints and conditions of the UAV routing problem are proposed, including modified simulated annealing method for UAV delivery \cite{dorling2016vehicle}, particle swarm optimization method for UAV routing with time window \cite{jiang2017method}, modified variable neighborhood search method for UAV routing in rescue mission \cite{mersheeva2012routing}, modified column generation algorithm for multi-UAV routing in complex missions \cite{mufalli2012simultaneous}, etc. We refer to \cite{chung2020optimization} for a systematical overview of the traditional optimization approaches for UAV routing. 

Recent advances in artificial intelligence brings new possibilities to combinatorial optimization. Vinyals et al. \cite{vinyals2015pointer} proposed the first seminal work of Pointer Networks that use a deep neural network to output the solution of combinatorial optimization problems. The model structure is similar to the sequence-to-sequence model for machine translation, i.e., mapping from the problem input sequence to the solution sequence. The model is trained by supervised learning, and is shown to be effective on small scale combinatorial optimization problems. 

Bello et al. \cite{bello2016neural} proposed to use the reinforcement learning method to train the Pointer Network. The cost of an instance is considered as the reward in the reinforcement learning process. Actor-Critic algorithm is used to optimize the model parameters. This overcomes the limitation of supervised learning \cite{vinyals2015pointer} that requires high-quality labelled train data. This model exhibits competitive performance with \cite{vinyals2015pointer} on small instances. It can additionally solve larger instances like 100-node TSP. 

Nazari et al. \cite{nazari2018deep} introduced a modified Pointer Network model with dynamic elements. By introducing the dynamic elements, this model can solve the capacity constrained VRP and the stochastic VRP. Moreover, it replaces the LSTM encoder of the Pointer Network by a simple embedded input. This reduces the model training time by 60\% with nearly identical solution quality on TSP instances.

Dai et al. \cite{dai2017learning} introduced a graph neural network model \emph{structure2vec} for combinatorial optimization. The node is inserted into the partial solution step by step, according to node scores computed by the \emph{structure2vec}. The node scores are considered as Q values, and the model is then trained via the Q-learning method. They apply this model on various combinatorial optimization problems like TSP, Minimum Vertex Cover and Maximum Cut problems. Mittal et al. \cite{mittal2019learning} followed the framework of \cite{dai2017learning}, but replaced \emph{structure2vec} with a better graph convolutional network (GCN) model. Their model is shown to  be better than \cite{dai2017learning} on large instances.

The authors in \cite{deudon2018learning, kool2018attention} followed the sequence-to-sequence framework, and replaced the encoder with the encoder of the Transformer model \cite{vaswani2017attention}, which is one of the most widely used model in the field of machine translation. Deudon et al. \cite{deudon2018learning} improved the solution found by the neural network by a simple local search. Kool et al. \cite{kool2018attention} modified the decoder of the model, and designed a greedy rollout baseline method to train the model. They apply this model on a variety of routing problems, such as TSP, VRP, and orienteering problem (OP). 

The authors in \cite{chen2019learning, lu2019learning} proposed to improve the search-based method for solving the routing problem by the means of deep reinforcement learning. Chen et al. \cite{chen2019learning} proposed a \emph{NeuRewriter} model that learns a policy to iteratively improve the solution until convergence. Lu et al. \cite{lu2019learning} learns to improve the solution iteratively with an improvement operator, selected from a collection of operators by a reinforcement learning model. Their models both outperform traditional solvers like Google OR-tools, however, as search-based approaches, they run much slower than the above deep learning approaches like \cite{kool2018attention}. For example, the model of \cite{chen2019learning} costs 24 minutes to solve the 100-node VRP. Moreover, Li et al. \cite{li2020deep, li2021} proposed to use deep reinforcement learning method to solve the multi-objective traveling salesman problem and the covering salesman problem, and obtained promising results.

\section{System model}\label{system}

This study considers a typical scenario where a rotary-wing UAV departs from the base station with a full power state, travels to each task point once, and returns to the base station in the end. The UAV has to fly back to the base station to get charged wirelessly once its State of Charge (SoC) decreases to a given level (20\% in this work). The task point and base station are distributed in a 2D  topology, i.e., located via the X-Y coordinates. The UAV is assumed to fly at a fixed height $H$. 

In this work, the UAV is charged by the wireless power transmission technology. It means that the UAV needs not to land and get charged by human. It is beneficial for the case when the working environment of the UAV is complicated, making it not safe or convenient for the UAV to land, which is especially common in military applications \cite{zeng2019energy}. One may need to wait, collect, charge, and fly the UAV if the wireless power transfer technology is not used. Human costs are effectively reduced, and a complete unmanned operation of the UAV can be achieved by applying the wireless power transfer technology.

In our study, the energy transfer process starts when the UAV reaches at a threshold horizontal distance $x_{start}$ to the base station. This distance is determined by the activation energy of the energy harvester on the UAV, that is, the minimum power the energy harvester requires to start charging the battery. Then, the UAV arrives at the top of the base station, and continues to get charged until the battery is full. 

Under this scenario, we introduce the UAV power consumption model and the energy charging model where the wireless transmission loss and the conversion loss from radio frequency to direct current (RF-DC) are considered.

\subsection{UAV Power Consumption Model}

The UAV is assumed to fly at a constant speed when traveling between the task points and the depot. For rotary-wing UAVs at a constant speed of $V$, the power consumption model is derived as \cite{yan2020uav}: 

\begin{equation}
P(V)=P_{0}\left(1+\frac{3 V^{2}}{U_{t i p}^{2}}\right)+P_{i}\left(\sqrt{1+\frac{V^{4}}{4 v_{0}^{4}}}-\frac{V^{2}}{2 v_{0}^{2}}\right)^{\frac{1}{2}}+\frac{d_{0} \rho s A V^{3}}{2}
\label{power_fly}
\end{equation}

where $P_0$ and $P_i$ are blade power and induced power:

\begin{equation}
P_{0}=\frac{\delta}{8} \rho s A \Omega^{3} R^{3}
\end{equation}

\begin{equation}
P_{i}=(1+k) \frac{W^{\frac{3}{2}}}{\sqrt{2 \rho A}}
\end{equation}
where $\rho$, $s$, $A$, $\Omega$, $R$, $W$ are the air density, rotor solidity, rotor disc area, blade angular velocity, rotor radius and UAV weight, respectively. $P_0$ and $P_i$ are constants if the UAV properties and the flying environment are invariable. $v_0$ represents the mean rotor induced velocity, and $U_{t i p}$ is the tip speed of the rotor blade. They are calculated as:

\begin{equation}
v_{0}=\sqrt{\frac{W}{2 \rho A}}
\end{equation}

\begin{equation}
U_{t i p} \triangleq \Omega R
\end{equation}

And $d_{0} \triangleq \frac{S_{F P}}{s A}$ is the fuselage drag ratio where $S_{F P}$ indicates the fuselage equivalent flat plate area. 

In this case, the power that a UAV consumes when flying at a speed $V$ can be calculated by \eref{power_fly} as $P(V)$. When the UAV is hovering, i.e., $V=0$, its power consumption is $P(0) = P_0+P_i$. Thus the energy required for flying and hovering for a rotary-wing UAV can be computed as $E_{V}=P(V) \cdot T_{\text {flying }}$ and $E_{hover}=P(0) \cdot T_{\text {hover }}$

\subsection{Wireless Charging Model}

When the base station starts to charge the UAV, a Line of Sight (LoS) link is established between the UAV and the base station. Thus, the free-space path loss (FSPL) model \cite{zeng2019energy} can be used to calculate the wireless transmission loss of the link:

\begin{equation}
P L(d)=20 \lg \{f\}+20 \lg \{d\}-147.55 d B
\end{equation}

where $f$ is the carrier frequency. And $d$ is the distance between the UAV and the base station. 

The RF-to-DC conversion loss is also considered in this work. A number of linear and non-linear models are studied for the RF-to-DC conversion loss. The linear model \cite{yan2020uav} is adopted in this work with a fixed conversion efficiency $\eta$:

\begin{equation}
P_{D C}=\eta \cdot P_{R F}
\end{equation}
where $P_{R F}$ is the input RF power to the energy harvester, and $P_{D C}$ is the converted DC power to the battery.

Denote the received power of the energy harvested at the UAV as $P_{charge}(d)$ and the distance between the UAV and the base station as $d$, one has:

\begin{equation}
P_{charge}(d)=P_{t}+G_{t}+G_{u a v}-P L(d)
\end{equation}

where $P_{t}$ is the transmitted power from the base station in dBW, $G_t$ is the gain of the transmitting antenna at the base station in dBi, and $G_{uav}$ is the receiving antenna gain at the UAV in dBi. $P L(d)$ is the transmission loss of the wireless channel between the base station and the UAV in dB. That is:

\begin{equation}
P_{charge}(d)=P_{t}+G_{t}+G_{u a v}-20 \lg \{f\} - 20 \lg \{d\}+147.55 d B
\end{equation}

It can be seen that the charging power changes according to the distance $d$. In this case, the charging process can be divided into two stages: the UAV is charged before it arrives at the top of the base station; the UAV is charged when hovering at the top of the base station. 

\subsubsection{Charging Model When Hovering}

During the second charging stage where the UAV is hovering at the height of $H$, the charging power is constant:

\begin{equation}
P_{\text {charge}}^{\text {hover}}=P_{t}+G_{t}+G_{u a v}-P L(H)
\end{equation}

Therefore, the received DC energy of the UAV when hovering can be calculated as:

\begin{equation}
E_{\text {charge}}^{\text {hover}}=\eta 10^{\frac{P_{\text {charge}}^{\text {hover}}}{10}} T^{\text {hover}}
\end{equation}

Note that the UAV still uses energy when hovering. The consumed energy is calculated as:
\begin{equation}
E_{\text {use}}^{\text {hover}}=P(0) \cdot T^{\text {hover}}
\end{equation}

Denote the required energy of the UAV as $E_{\text {hover}}$, the charging time when hover can be calculated as:
\begin{equation}
T^{\text {hover}}=\frac{E_{\text {hover}}}{\eta10 \frac{P_{\text {uav-r }}}{10}-P(0)}
\end{equation}

\subsubsection{Charging Model When Flying}

The base station starts to charge the UAV when the UAV is flying to the base station. Note that the energy harvester of the UAV has a minimum power threshold $P_\epsilon$ to activate its energy charging. Denote the threshold location of the UAV as $x_{\text{start}}$, at which the UAV starts to charge, one has:

\begin{equation}
P_{t}+G_{t}+G_{u a v}-20 \lg \{f\} - 20 \lg \{d_{\text{start}}\}+147.55 d B \geqslant P_\epsilon dB
\end{equation}
where $d_{\text{start}} = \sqrt{\left(x_{bs}-x_{\text {start }}\right)^{2}+\Delta H^{2}}$ is the distance between the base station and the threshold location of the UAV. Therefore, the location at which the UAV starts to charge can be calculated as:

\begin{equation}
x_{\text {start }} = x_{1}-\sqrt{10^{\frac{P_{t}+G_{t}+G_{u a v}-20 \lg \{f\} +147.55 - P_\epsilon}{10}}-\Delta H^{2}}.
\end{equation}

\begin{figure}[!t]
\centering
\includegraphics[width=3.5in]{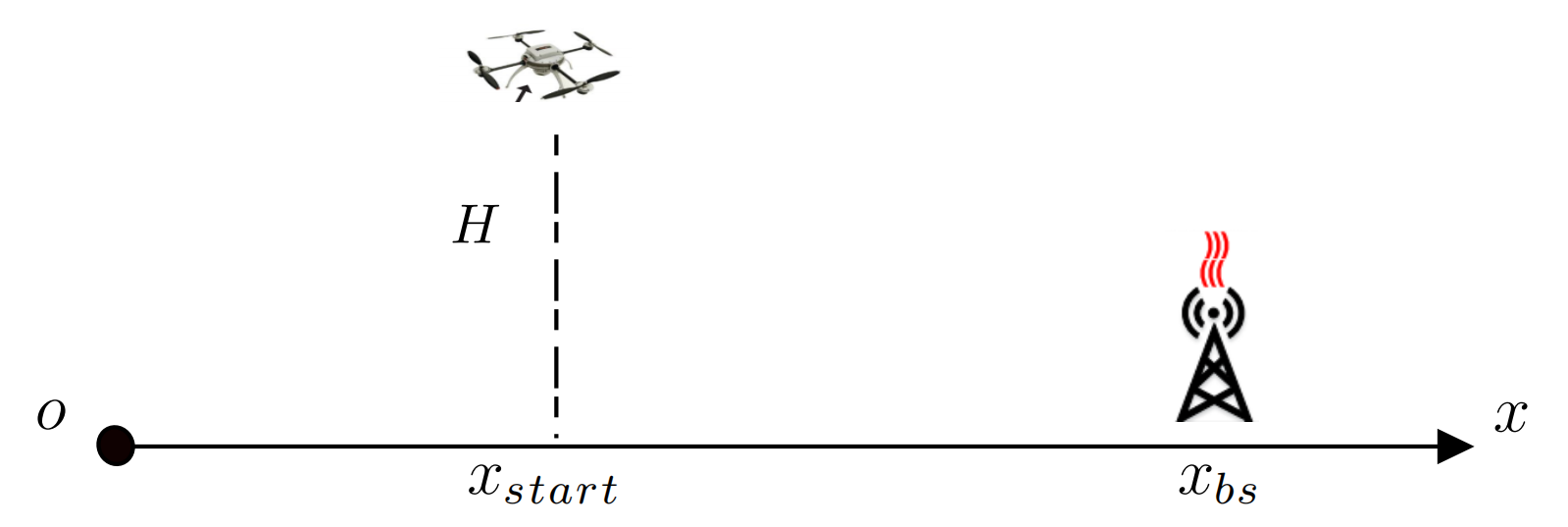}
\caption{Wireless charging process. The base station starts to charge the UAV when they are close each other. The threshold distance is represented by $x_{start}$.}
\label{charge}
\end{figure}

This charging period is visualized in \fref{charge}.

Assuming that the charging period starts at time $t=0$ and the speed of the UAV is $V$, at time $t$, the distance between the UAV and the station is:

\begin{equation}
d_t=\sqrt{\left(x_{s t a r t}+V t-x_{bs}\right)^{2}+\Delta H^{2}}
\end{equation}

Then, the charging power at time $t$ is:

\begin{equation}
P_{charge}^t=\Omega-20 \lg \left(d_t\right)
\end{equation}
where $\Omega=P_{t}+G_{t}+G_{u a v}-20 \lg (f)+147.55$

Denote the time that the UAV spends when flying from $x_{\text {start }}$ to the top of the base station as $T_{\text {charge }}^{fly} = (x_{bs} - x_{start})/V$, the total energy that the UAV receives during this charging period is \cite{yan2020uav}:

\begin{equation}
\begin{aligned}
E_{fly}=& \eta \int_{0}^{T_{\text {charge }}^{fly}} 10^{ \frac{P_{charge}^t}{10} }d t \\
=& \frac{2 \eta 10^{\frac{\Omega}{10}}}{\sqrt{4 A_{2} C_{2}-B_{2}^{2}}} \arctan \frac{B_{2}+2 C_{2} T_{\text {charge }}^{fly}}{\sqrt{4 A_{2} C_{2}-B_{2}^{2}}} \\
&-\frac{2 \eta 10^{\frac{\Omega}{10}}}{\sqrt{4 A_{2} C_{2}-B_{2}^{2}}} \arctan \frac{B_{2}}{\sqrt{4 A_{2} C_{2}-B_{2}^{2}}}
\end{aligned}
\end{equation}
where $A_{2}=\left(x_{\text {start }}-x_{bs}\right)^{2}+\Delta H^{2}$, $B_{2}=2 V\left(x_{\text {start }}-x_{bs}\right)$, and $C_2 = V^2$.

Denote the remaining energy of the UAV when it arrives at $x_{\text {start }}$ as $E_{r}$, the time  that the UAV requires to hover over the base station can be calculated as:

\begin{equation}
T^{\text {hover}} = \frac{E_{full} - E_r - E_{fly}}{\eta10 \frac{P_{\text {uav-r }}}{10}-P(0)}
\end{equation}

\subsection{Optimization Model}

The objective of the problem is to minimize the total mission time of the UAV, including the flying time and the charging time. Let $d_{ij}$ be the distance from node $i$ to node $j$ and node with index 0 be the base station. Assuming that the UAV returns to the base station for $K$ times, i.e., the solution consists of $K$ Hamiltonian cycles. Let $x_{ijk}$ be the decision variable, which is binary that has value 1 if the arc going from $i$ to $j$ is considered as part of the Hamiltonian cycle $k$ in the solution. Therefore, The mathematical formulation for the problem is:

\begin{equation}
\min \sum_{i, j=0 \atop i \neq j}^{N}\sum_{k=1}^K d_{i j} x_{i jk} / V + \sum_{i=0}^K T^{\text {hover}}_i
\end{equation}

subject to 

\begin{equation}
\sum_{k=1}^{K} \sum_{j=1}^{N} x_{0 j k}=\sum_{k=1}^{K} \sum_{j=1}^{N} x_{j 0 k}=K
\label{c1}
\end{equation}
\begin{equation}
\sum_{i=1}^{N} x_{i j k}=\sum_{i=1}^{N} x_{j ik},(k \in K, \forall j=1,2, \ldots, N)
\label{c2}
\end{equation}
\begin{equation}
\sum_{i=1}^{K} \sum_{i=0}^{N} x_{ij k}=1,(j=1,2, \ldots, N)
\label{c3}
\end{equation}
\begin{equation}
P(V) \sum_{i=0}^{N} \sum_{j=0}^{N} \mathrm{x}_{i j k} d_{i j}/V \leq (1-SOC_{min})E_{full}, (k \in K)
\label{c4}
\end{equation}

Constraints \eref{c1} states that the number of the UAV entering into the base station is the same as the number of those leaving. Constraints \eref{c2} says that the number of the UAV entering into each node is the same as the number of those leaving. Constraints \eref{c3} states that each node is visited exactly once. Constraints \eref{c4} is the energy constraint, where the SOC of the UAV should be larger than $SOC_{min}$ when it arrives at the top of the base station. 

\section{Deep Neural Network Model}\label{smodel}

We use the sequence-to-sequence model infrastructure similar to \cite{kool2018attention} to solve the proposed problem. The problem is modeled as a sequence-to-sequence task: mapping from the location sequence of the nodes to the node index sequence. In this way, we can leverage the recent advances in machine translation field, where the sequence-to-sequence models are widely studied. 

\begin{figure*}[!t]
\centering
\includegraphics[width=4.5in]{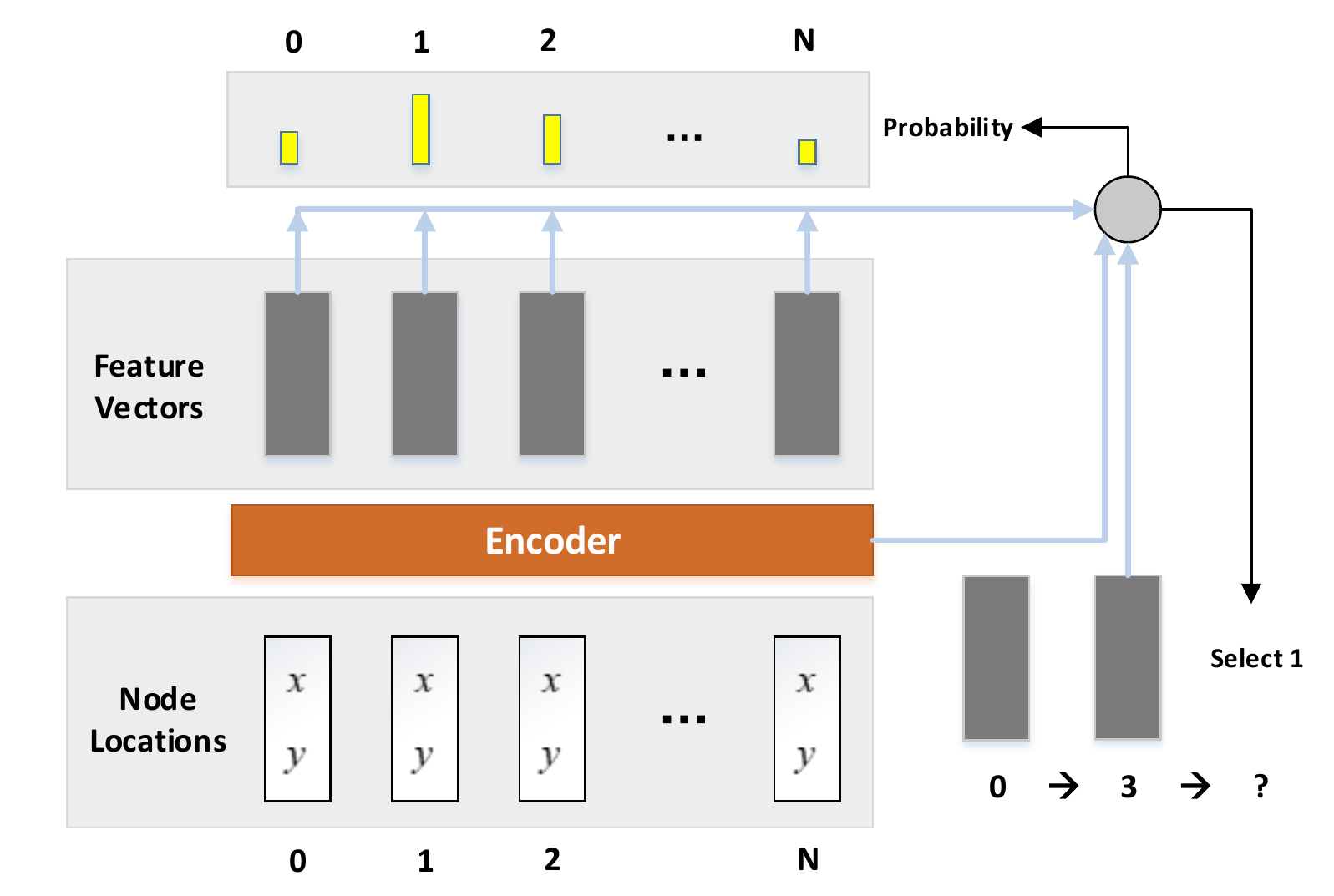}
\caption{Structure of the model. Encoder is shown in the left. The decoding process of the decoder is shown in the right. The yellow columns represent the probabilities of selecting each node at the current decoding step, computed by the decoder.}
\label{model}
\end{figure*}

The model consists of two parts: encoder and decoder. Encoder computes the $d_h$-dimensional feature vector $\mathbf{h}_{i}$ for all nodes given their initial features, i.e., two-dimensional location coordinates. In this work, $d_h$ is set to 128. Decoder builds the solution autoregressively, i.e., outputs the tour one node at a time. In decoder, the node with the largest probability is chosen as the next node to visit. The probability of node $i$ is calculated according to its feature vector $\mathbf{h}_{i}$ and the current decoder context, which infers the current state of the solution. The model architecture is shown in \fref{model}.

\subsection{Encoder}

In encoder, the multi-head attention (MHA) similar to \cite{kool2018attention} is adopted to compute the feature vector $\mathbf{h}_{i}$ for each node $i$. Multi-head attention is proposed in the Transformer model \cite{vaswani2017attention}, which is probably the most widely used method for machine translation tasks. It can extract richer information than other methods. 

First, given the two-dimensional locations $\mathbf{x}$ of all nodes, the initial $d_h$-dimensional node embedding is calculated as:

\begin{equation}
\mathbf{h}_{i}^{(0)}=W^{\mathbf{x}} \mathbf{h}_{i}^{(0)}+\mathbf{b}^{\mathbf{x}}.
\end{equation}
where $W^{\mathbf{x}}$ and $\mathbf{b}^{\mathbf{x}}$ are the learnable parameters of the model. $\mathbf{h}^{(0)}$ serves as the initial feature vector of the nodes. MHA converts $\mathbf{h}^{(0)}$ to the final node feature vectors.

Self-attention, sometimes called intra-attention is the basic operator of MHA. It computes the attention value of each node to all other nodes in the input sequence. Therefore, the obtained attention value (feature vector) of each node stores not only its own information, but also its relationship with other nodes. For node $i$, its attention value is computed by its \emph{query} and the \emph{key-value} pairs of all the nodes. Each node has a \emph{query}, a \emph{key} and a \emph{value}. All of them are vectors and they all come from the input sequence. In specific, \emph{query}, \emph{key} and \emph{value} of node $i$ are linear projections of the its initial node embedding $\mathbf{h}_{i}^{(0)}$:

\begin{equation}
\mathbf{q}_{i}=W^{Q} \mathbf{h}_{i}^{(0)}, \quad \mathbf{k}_{i}=W^{K} \mathbf{h}_{i}^{(0)}, \quad \mathbf{v}_{i}=W^{V} \mathbf{h}_{i}^{(0)}.
\end{equation}

For all of the nodes, their \emph{queries}, \emph{keys} and \emph{values} are:
\begin{equation}
\mathbf{Q}=W^{Q} \mathbf{X}, \quad \mathbf{K}=W^{K} \mathbf{X}, \quad \mathbf{V}=W^{V} \mathbf{X}.
\label{qkv}
\end{equation}

Assume there are $N$ nodes in the input sequence. For node $i$, its attention value is the weighted sum of the $N$ \emph{values}, where the weight is computed by a compatibility function of its \emph{query} with the $N$ \emph{keys} of all the nodes. The weight is computed as $\mathbf{q}_i\mathbf{K}^T$. Therefore, the attention value of node $i$ is $\mathbf{q}_i\mathbf{K}^T\mathbf{V}$. The \texttt{softmax} function is used to normalize the weight. Then we compute the matrix of the attention values of all nodes as:

\begin{equation}
\text { Attention }(Q, K, V)=\operatorname{\texttt{softmax}}\left(\frac{Q K^{T}}{\sqrt{d_{h}}}\right) V
\end{equation}

where $1/{d_h}$ is the scaling factor. MHA splits $\mathbf{h}^{(0)}$ into $M $ sub-vectors, and conducts the self-attention $M$ times to obtain $M$ attention values with $d_h/M$ dimensions. The $M$ attention values are then concatenated and linearly projected into the final $d_h$-dimensional values. MHA allows the model to convey richer information of the node features.

\begin{figure}[!t]
\centering
\includegraphics[width=3in]{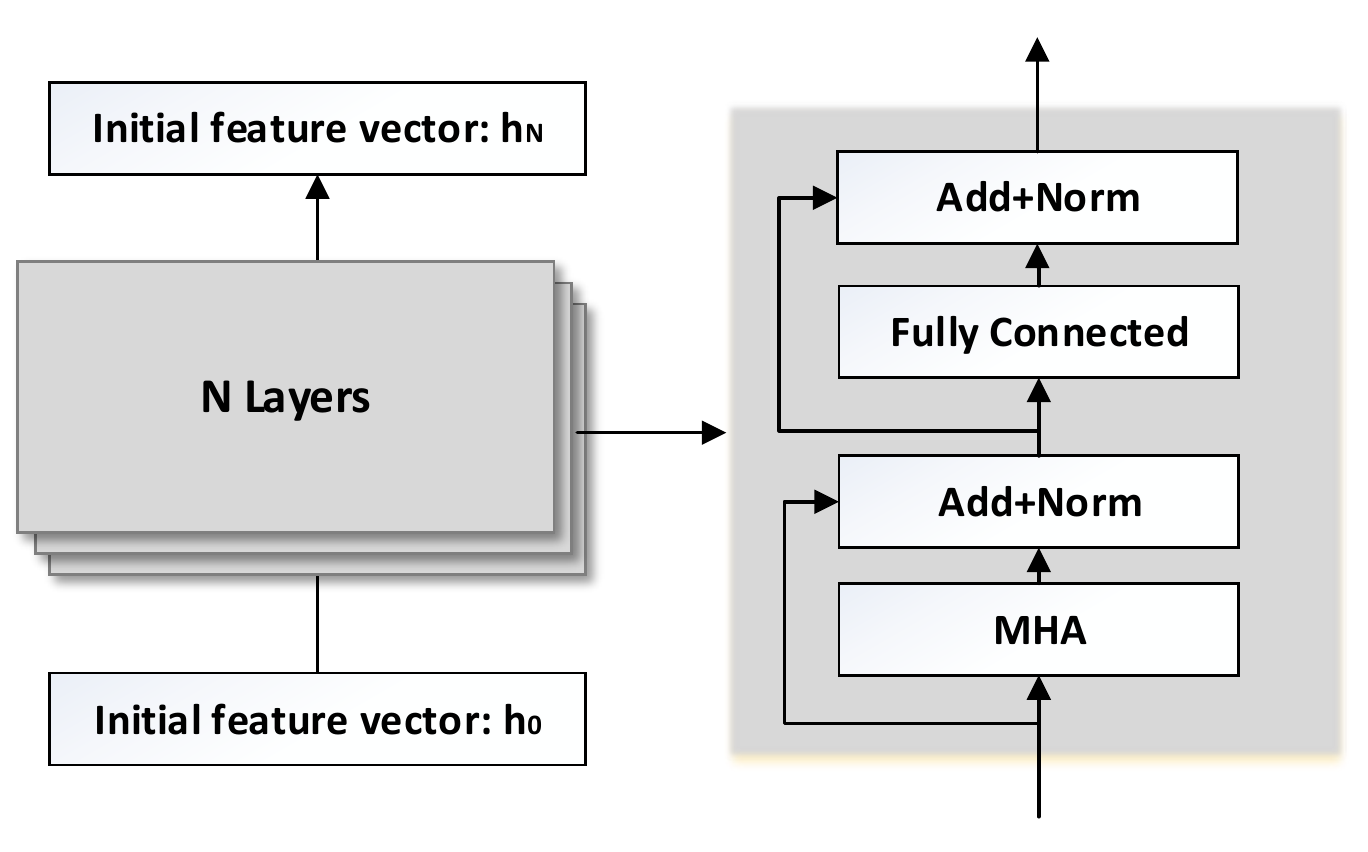}
\caption{Structure of the encoder.}
\label{encoder}
\end{figure}

The architecture of the encoder is shown in \fref{encoder}, which is composed of $N$ sequentially connected attention layers. Each layer consists of a MHA layer and a fully connected layer. The residual connection and layer normalization are also used to process the layer output. The output of layer ${\ell}$ is computed as:
\begin{equation}
\begin{aligned} \\\mathbf{h}_{tmp} &=Norm^{\ell}\left(\mathbf{h}^{(\ell-1)}+\mathrm{MHA}^{\ell}\left(\mathbf{h}^{(\ell-1)}\right)\right) \\ 
\mathbf{h}^{(\ell)} &=Norm^{\ell}
\left(\mathbf{h}_{tmp}+\mathrm{FF}^{\ell}\left(\mathbf{h}_{tmp}\right)\right) \end{aligned}.
\label{eq:mha}
\end{equation}
where $\mathbf{h}^{(\ell-1)}$ is the output of the previous layer, $\mathrm{FF}$ is the fully connected layer, and $\mathrm{MHA}$ is the multi-head attention layer. By processing the initial node embedding $\mathbf{h}^{(0)}$ via the $N$ layers according to \eref{eq:mha}, we can obtain the final feature vectors $\mathbf{h}^{(N)}$ for all the nodes.

\subsection{Decoder}
Decoder selects the node step by step. At each decoding step $t$, the node $\pi_t$ with the largest probability is selected. The probabilities of the nodes are computed by the feature vectors $\mathbf{h}^{(N)}$ of the nodes and the current decoding context. The decoding context at step $t$ represents the current state of the partial solution, including the partial tour generated at time $t^{\prime}<t$ and the current SOC of the UAV.

The SOC is initialized as $SOC_t = 1$ every time the UAV departs from the base station, after which it is updated as follows:

\begin{equation}
{SOC}_{t+1}=\left\{\begin{array}{ll}
{SOC}_{t}-P(V)d(\pi_{t}, \pi_{t-1})/V/E_{full} & \pi_{t} \neq 0 \\
1 & \pi_{t}=0
\end{array}\right.
\end{equation}
where the node with the index $\pi_t=0$ represents the base station, and $d(*,*)$ represents the distance between the two nodes.

The decoding context at step $t$ is designed as:

\begin{equation}
\mathbf{d}_t=\left\{\begin{array}{ll}
{\left[\overline{\mathbf{h}}^{(N)}, \mathbf{h}_{\pi_{t-1}}^{(N)}, SOC_{t}\right]} & t>1 \\
\left[\overline{\mathbf{h}}^{(N)}, \mathbf{h}_{0}^{(N)}, SOC_{t}\right] & t=1
\end{array}\right.
\end{equation}
where $\mathbf{h}_{\pi_{t-1}}^{(N)}$ is the feature vector of the node selected at step $t-1$. $\overline{\mathbf{h}}^{(N)}$ is the mean value of the feature vectors ${\mathbf{h}}^{(N)}$ of all the nodes, enabling the model understand the overall state of the problem. 

The decoding context can be seen as the \emph{query}. Meanwhile, each node owns a $key$ represented by ${\mathbf{h}}^{(N)}$. Among all $key_i, i=1 \ldots N$, the one that most matches the \emph{query} can be selected as the next node to visit. This process is realized using the attention method similar to the encoder. First we compute the \emph{query} and \emph{key}:

\begin{equation}
\mathbf{q}_{t}=W^{Q'} \mathbf{d}_t, \quad \mathbf{k}_{i}=W^{K'} \mathbf{h}_{i}^{(N)}
\end{equation}

Then the probabilities for selecting the nodes can be computed as:

\begin{equation}
u_{i}=\left\{\begin{array}{ll}
C \cdot \tanh \left(\frac{\mathbf{q}_{t}^{T} \mathbf{k}_{i}}{\sqrt{d_{h}}}\right) & \text { if } i  \text { is not masked }\\
-\infty & \text { otherwise. }
\end{array}\right.
\end{equation}
where $C$ is used to clip the result. The \texttt{softmax} function is used to normalize $u_{i}$ to obtain the final probabilities.
Here, a node is masked if it cannot be selected, including the following situations. First, node $i>0$ is masked if it has been already visited at time $t'<t$. Second, node $i=0$, i.e., the base station, is masked if it is selected at the last step $t-1$, since the UAV should leave the base station if it has been charged. Moreover, node $i>0$ is masked if:
\begin{equation}
P(V)\left( d(\pi_{t-1},i)+d(i,0) \right)/V > (SOC_{t-1}-SOC_{min})E_{full}
\end{equation}
That is, the remaining electricity should be enough for the UAV to travel to node $i$ and return to the base station. 

The decoding procedure is visualized in \fref{model}. In \fref{model}, nodes 0 and 3 have been selected, and the task is to determine the next node. The probabilities are computed by the current decoder context and the feature vectors from the encoder. Node 1 with the largest probability is thus selected. This decoding procedure is looped until all the nodes are visited and the UAV returns to the base station.

\section{Deep Reinforcement Learning}\label{train}

We use deep reinforcement learning (DRL) to train the proposed deep neural network model. Formally, given a problem instance $s$, the model outputs the probability distribution $p_{\boldsymbol{\theta}}(\boldsymbol{\pi} | s)$ to produce the solution: 

\begin{equation}
p_{\theta}(\boldsymbol{\pi} | s)=\prod_{t=1}^{k} p_{\boldsymbol{\theta}}\left(\pi_{t} | s, \boldsymbol{\pi}_{1 \sim t-1}\right), k \leq N.
\end{equation}
That is, at each step $t$, the model outputs the probability of selecting $\pi_{t}$ given the instance $s$ and the current partial solution $\boldsymbol{\pi}_{1 \sim t-1}$. The model is parameterized by the neural network parameters $\boldsymbol{\theta}$, like $W^{Q}$ and $W^{K}$. The solution can be produced by sampling from $p_{\boldsymbol{\theta}}(\boldsymbol{\pi} | s)$, e.g., greedily selecting the node with the largest probability. 

We aim to minimize the total time $T(\pi)$ consumed by the UAV. Thus, the model parameters $\boldsymbol{\theta}$ are optimized by minimizing $T(\pi)$. There are two ways to optimize the neural network model parameters: supervised learning and unsupervised learning like reinforcement learning. Supervised learning requires generating a large number of problem instances with their optimized solutions as labels, which is intractable and not practical. Thereby, we use the reinforcement learning method to train the model. As the total time $T(\pi)$ can only be obtained at the end of an episode, we use the REINFORCE algorithm, a kind of Monte Carlo policy gradient algorithm, to train the model parameters in this study. Given a problem instance $s$ and the probability distribution $p_{\boldsymbol{\theta}}(\boldsymbol{\pi} | s)$ output by the model, model parameters are optimized by policy gradient as follows:

\begin{equation}
\begin{aligned}
\nabla_{\boldsymbol{\theta}} \mathcal{L}(\boldsymbol{\theta}) & = \mathbf{E}
_{p_{\boldsymbol{\theta}}(\boldsymbol{\pi} | s)}
\left[ \nabla \log p_{\boldsymbol{\theta}}(\boldsymbol{\pi} | s)T(\boldsymbol{\pi})\right]
\\
\boldsymbol{\theta} &  \leftarrow \boldsymbol{\theta} +\nabla_{\boldsymbol{\theta}} \mathcal{L}(\boldsymbol{\theta}).
\end{aligned}
\label{PG}
\end{equation}

It is common to introduce a baseline function $b(s)$ to improve \eref{PG}:

\begin{equation}
\nabla_{\boldsymbol{\theta}} \mathcal{L}(\boldsymbol{\theta}) = \mathbf{E}
_{p_{\boldsymbol{\theta}}(\boldsymbol{\pi} | s)}
\left[ \nabla \log p_{\boldsymbol{\theta}}(\boldsymbol{\pi} | s)(T(\boldsymbol{\pi})-b(s))\right].
\label{PG2}
\end{equation}
 where $b(s)$ represents the average performance: if $T(\boldsymbol{\pi})-b(s) < 0$, the policy $p_{\boldsymbol{\theta}}(\boldsymbol{\pi} | s)$ is encouraged since it performs better than the average performance $b(s)$, and vice versa. 
 
In this work, we apply the greedy rollout baseline as $b(s)$, which is reported in  \cite{kool2018attention} that can effectively improve the training performance. In specific, the best performing model so far during training is stored as the baseline policy $\boldsymbol{\theta}^*$. $p_{\boldsymbol{\theta}^*}(\boldsymbol{\pi} | s)$ can be obtained by inputting $s$ into the baseline model $\boldsymbol{\theta}^*$. $b(s)$ is therefore computed as the total time $T(\boldsymbol{\pi}^*)$ of the solution $\boldsymbol{\pi}^*$, which is produced by greedily sampling from $p_{\boldsymbol{\theta}^*}(\boldsymbol{\pi} | s)$. In this way, $b(s)$ can represent the baseline performance, and guide the direction of the optimization. 

The training algorithm is outlined in Algorithm \ref{rl}. We first generate a number of instances $s$ randomly, i.e., a set of node locations. Then, the model $\boldsymbol{\theta}$ is used to produce the solution $\boldsymbol{\pi}$, and the corresponding $T(\boldsymbol{\pi})$ can be obtained. $\boldsymbol{\theta}$ can be updated according to \eref{PG2} to minimize $T(\boldsymbol{\pi})$. 

\begin{algorithm}
\caption{REINFORCE algorithm}\label{rl}
\KwData{Number of epochs $n_e$, number of steps per epoch $n_s$, batch size $B$.}
\KwResult{Optimal model parameters $\boldsymbol{\theta}$}
Initialize parameters of the main model $\boldsymbol{\theta}$ and the baseline model$\boldsymbol{\theta}$\;
\While{$epoch \leftarrow 1:n_e$}
{
\While{$step \leftarrow 1:n_s$}
{
generate $B$ instances $s_{i}$\;

$\boldsymbol{\pi}_{i} \leftarrow  p_{\boldsymbol{\theta}}(s_{i}).$ Run the main model and sample the solution randomly\;

$\boldsymbol{\pi}_{i}^* \leftarrow  p_{\boldsymbol{\theta}^*}(s_{i}).$ Run the baseline model and sample the solution greedily\;

 $\nabla \mathcal{L} \leftarrow \sum_{i=1}^{B}\left(T\left(\boldsymbol{\pi}_{i}\right)-T\left(\boldsymbol{\pi}_{i}^{*}\right)\right) \nabla_{\boldsymbol{\theta}} \log p_{\boldsymbol{\theta}}\left(\boldsymbol{\pi}_{i}\right) $ \;
 
$\boldsymbol{\theta}  \leftarrow \boldsymbol{\theta} +\nabla_{\boldsymbol{\theta}} \mathcal{L}(\boldsymbol{\theta}).$ Update parameters $\boldsymbol{\theta}$\;
}
}
\end{algorithm}

\section{Experiment}\label{exp}

\subsection{Experimental Settings}
Physical parameters of the UAV are listed in \tref{para}, which refers to \cite{yan2020uav}. In this study, the speed of the UAV is set to $10m/s$. Therefore, the power of the UAV when flying is $P(V)=62.49W$ according to \eref{power_fly}, and the power when hovering is $P(0)=82.46W$. $SOC_{min}$ is set to $20\%$ to ensure the safety of the UAV. The capacity of the UAV battery is $3830mAh$ ($43.8Wh$) and its voltage is $11.4V$. For the parameters of the wireless transmission device, we set $P_t=58$dBW, $G_t=58$dBi \cite{yan2020uav} and $f=915MHz$. In addition, the RF-to-DC conversion efficiency is $\eta=0.6$ \cite{yan2020uav}. And the threshold charging power of the UAV is set to $P_\epsilon=17$dBW. 

\begin{table}
\caption{Physical parameters of the UAV that refer to \cite{yan2020uav}.}
\begin{tabular}{c|c|c}
\hline 
Notation& Physical meaning & Value \\ 
\hline 
$m$ & Airframe mass in kg & 1.0 \\ 
\hline 
$W$ & Aircraft weight in Newton & 9.8 \\ 
\hline 
$\rho$ & Air density in $kg/m^3$& 1.225 \\ 
\hline 
$b$ & Number of blades & 4 \\ 
\hline 
$R$ & Rotor radius in $m$ & 0.25 \\ 
\hline 
$A$ & Rotor disc area in $m^2$ & 0.19634 \\ 
\hline 
$c$ & Blade length & 0.0196 \\ 
\hline 
$s$ & Rotor solidity & 0.0998 \\ 
\hline 
$\delta$ & Profile drag coefficient & 0.012 \\ 
\hline 
$\omega$ & Blade angular velocity in radians/s & 400 \\ 
\hline 
$k$ & Incremental correction factor to induced power & 0.05 \\ 
\hline 
$U_{tip}$ & Tip speed of the rotor blade & 100.0 \\ 
\hline 
$v_0$ & Mean rotor induced velocity & 4.5135 \\ 
\hline 
$S_{FP}$ & Fuselage equivalent flat plate area in $m^2$ & 0.0079 \\ 
\hline
$d_0$ & Fuselage drag ratio & 0.4030 \\ 
\hline 
$P_0$ & Blade power & 36.01 \\ 
\hline 
$P_i $ & Induced power & 46.44 \\ 
\hline 
\end{tabular} 
\label{para}
\end{table}

Hyperparameters of the proposed method are shown in \tref{para_m}. The dimension of the layers in the neural network model is all set to 128. 320,000 training instances are randomly generated for each epoch. Given the probability distribution $p_{\boldsymbol{\theta}}(\boldsymbol{\pi} | s)$ obtained by the trained model, three strategies are used to produce the solution:

\begin{enumerate}
\item Greedy. The node is selected greedily according the probability $p_{\boldsymbol{\theta}}(\boldsymbol{\pi} | s)$ at each decoding step. Only one solution is produced in this way.

\item Sample. The node is randomly sampled from the probability $p_{\boldsymbol{\theta}}(\boldsymbol{\pi} | s)$ at each decoding step. We produce $N_{sample}=1280$ solutions in this way, and select the best one as the output solution.

\item Beam search (BS). We apply the beam search strategy to produce the solution with beam width $N_{beam}=1000$. This method uses breadth-first search to build its search tree. However, it only reserves $N_{beam}$ best partial solutions during search. Only these states are expanded next. When $N_{beam}$ is infinite, beam search is the breadth-first search. 

\end{enumerate}

\begin{table}
\caption{Hyperparameters Configurations of the Model}
\begin{tabular}{cc|cc}
\hline 
HyperParameters & Value & HyperParameters & Value \\ 
\hline 
Batch size $B$ & 256 & Dimension of the network $d_h$ & 128 \\ 
\hline 
Epoch size $n_e$ & 100 & No. of heads $M$ & 8 \\ 
\hline 
Steps per epoch & 1250 & Attention layers & 3 \\ 
\hline 
Optimizer & Adam & Learning rate & 1e-4 \\ 
\hline 
\end{tabular} 
\label{para_m}
\end{table}

The proposed deep learning method is compared with the following benchmark approaches:

\begin{enumerate}
\item Google OR-tools \cite{ortools}. A well-known and high performing solver for solving combinatorial optimization problems with various meta-heuristics. We model the proposed problem in OR-tools using the vehicle routing problem solver template with default parameters.

\item The Clarke-Wright savings heuristic (CW) \cite{clarke1964scheduling}. It is one of the best-known heuristics especially designed for the routing problem. Algorithm parameters follow this open source repository\footnote{https://github.com/ishelo/VRP-CW}.

\item Particle swarm optimization algorithm (PSO) \cite{marinakis2019multi}.
\item Ant colony optimization algorithm (ACO) \cite{zhang2019hybrid}. 
\item Adaptive Large Neighborhood Search (ALNS) \cite{azi2014adaptive}. Algorithm parameters of 3), 4) and 5) all follow this open source repository\footnote{https://github.com/PariseC/Algorithms\_for\_solving\_VRP}.

\end{enumerate}

Note that a large number of other approaches can solve the proposed routing problem. But we did not further include these methods as competitors, because Google OR-tools can outperform most of the current approaches and won plenty of gold medals in various optimization competitions, like the MiniZinc Challenge 2020 and 2021. Therefore, we mainly take the Google OR-tools and some representative heuristic and local search algorithms as baselines to evaluate our method. 

All algorithms are implemented in Python to make the comparison fair. They are evaluated on the same machine with one GTX 2080Ti GPU and Intel 64GB 16-Core i7-9800X CPU. 100 randomly generated instances are taken as test set. And we record the average cost and running time of the compared approaches as performance indicators.

\subsection{Results and Discussion}

\begin{figure*}[htbp]
\centering
\subfigure[50-node]{\includegraphics[width=2.2in]{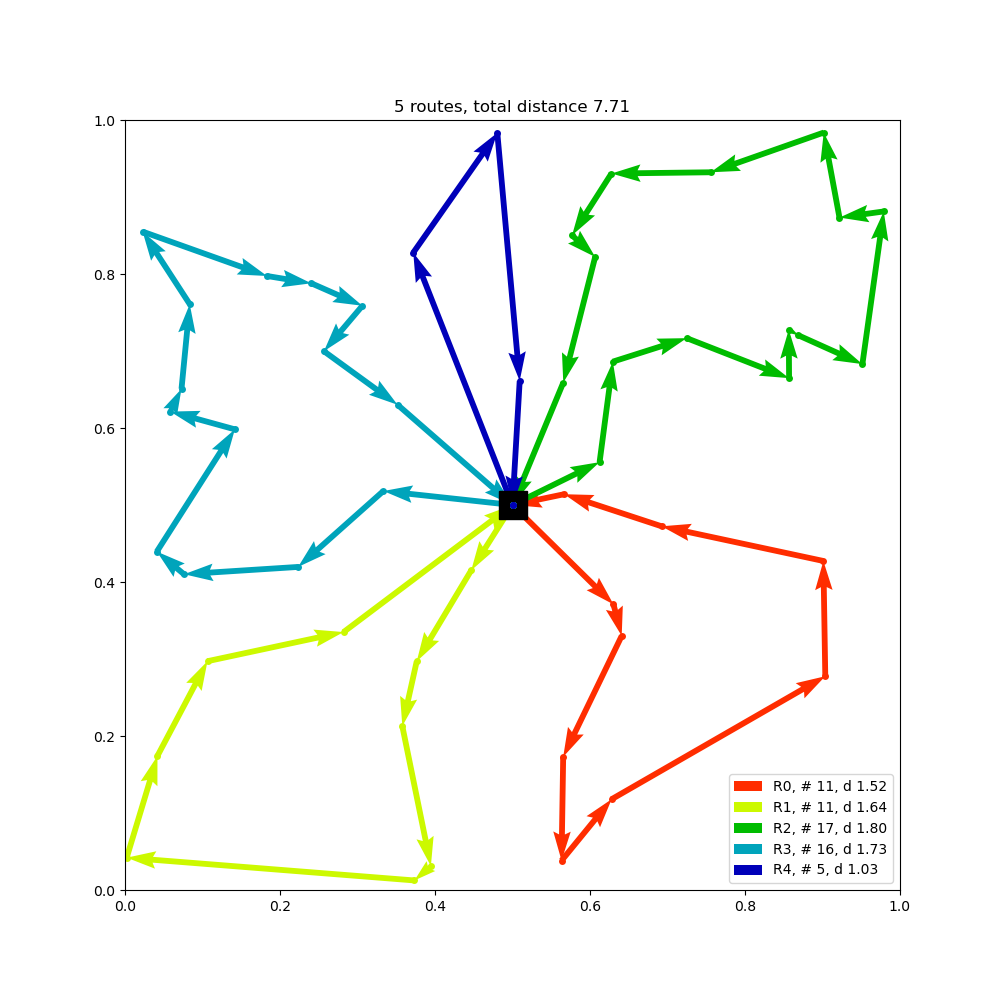}}
\hfil
\subfigure[100-node]{\includegraphics[width=2.2in]{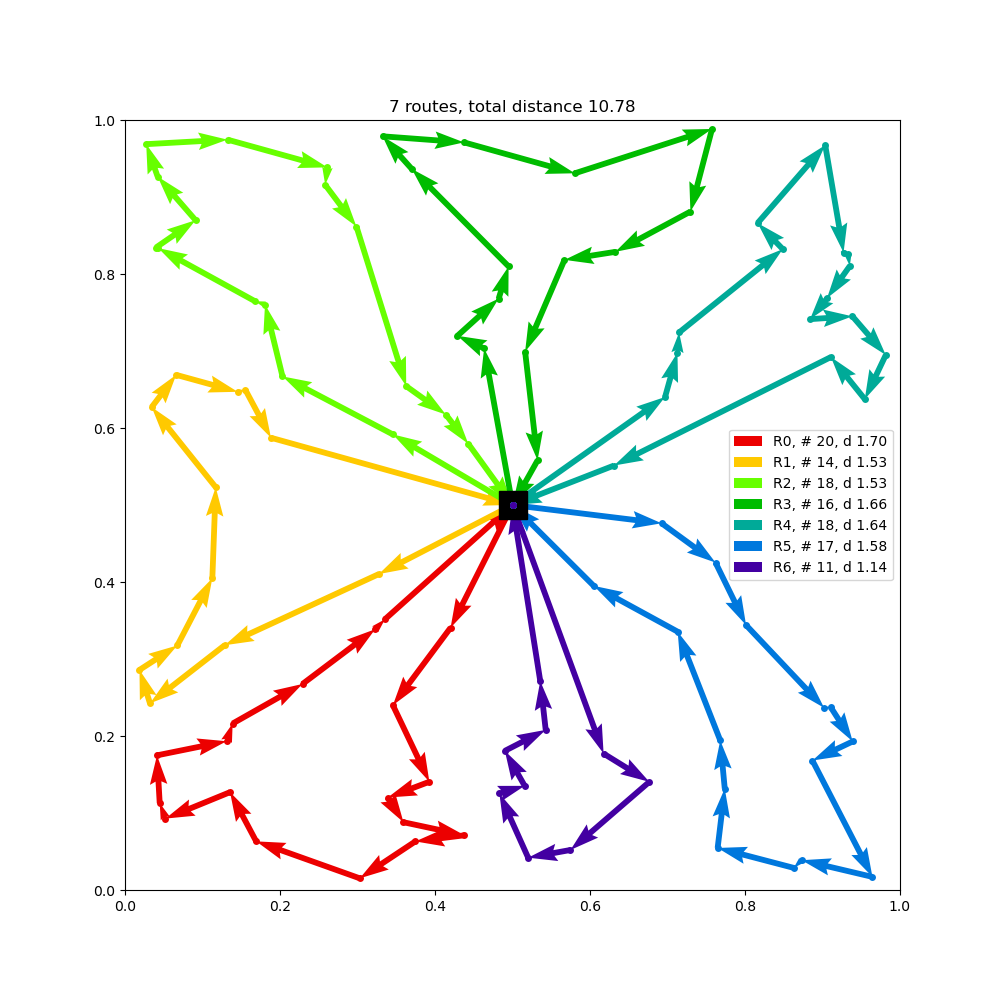}}
\hfil
\subfigure[50-node]{\includegraphics[width=2.2in]{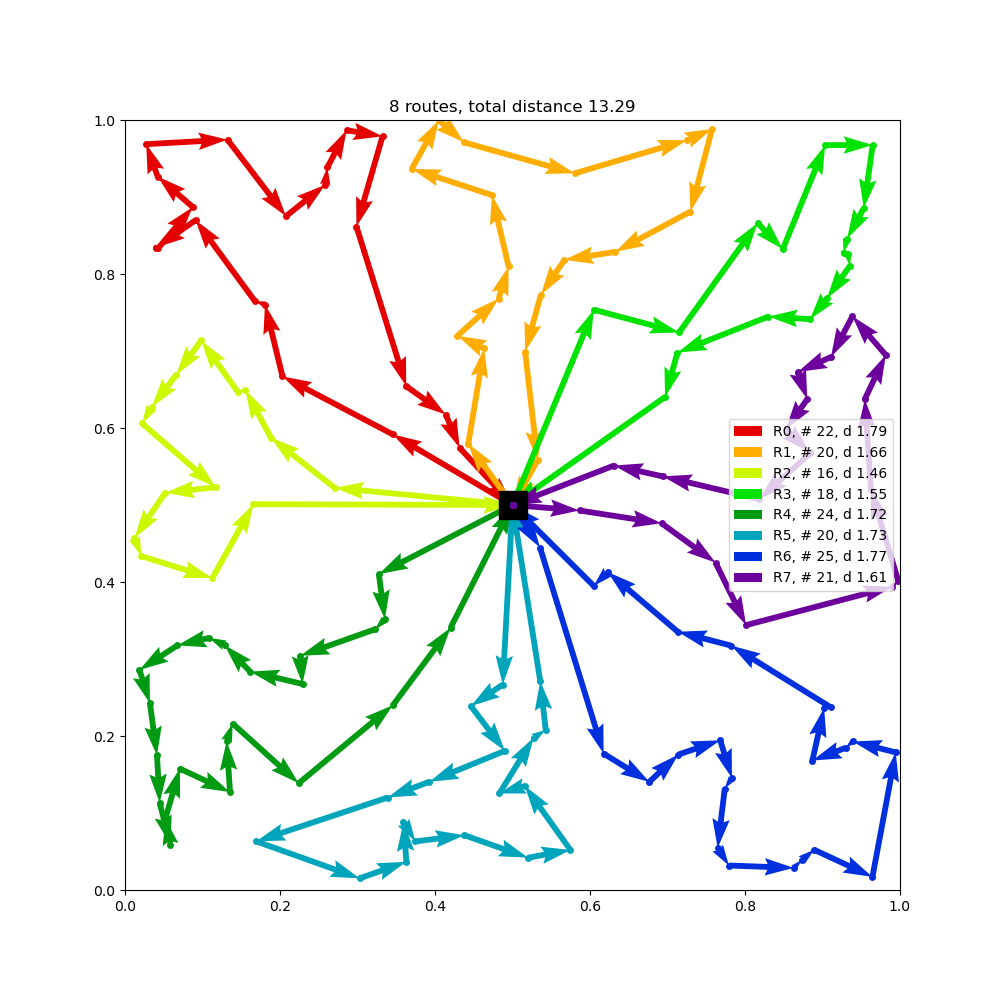}}
\hfil
\subfigure[100-node]{\includegraphics[width=2.2in]{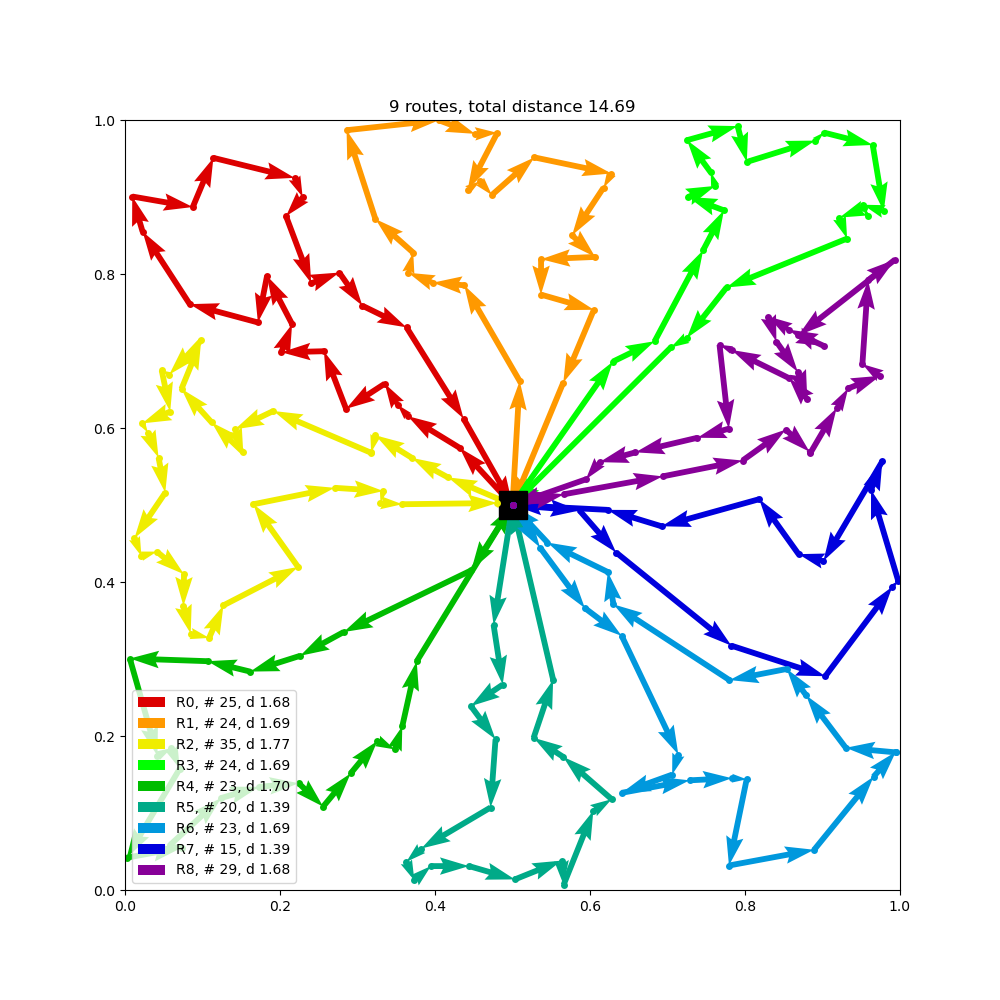}}
\caption{Solutions found by the proposed method on instances with different sizes.}
\label{fig:v}
\end{figure*}

\fref{fig:v} shows the solutions found by the proposed method on 50-, 100-, 150-, 200-node instances, using sampling strategy. It is observed that the proposed method can effectively solve the UAV routing problem with energy constraint. 

\tref{tab:small} shows the numerical results of the compared approaches on small scale instances with 20, 50 and 100 nodes. The cost is the average total time the UAV consumed in hour. The average algorithm run-time is also recorded. Results of our method with greedy, sample and beam search strategy are presented. The optimality gap between our method and the best performing method is listed. We also give the speed that our method improves over Google OR-Tools.

We observe that using sample and beam search strategy can improve the solution but increasing the run-time in the meantime. But the run-time is always within one second.

It can be seen that, the proposed deep learning method outperforms Google OR-Tools in terms of both optimality and solving speed on 20- and 50-node instances. On 100-node instances, the proposed method performs slightly worse (0.7\%) than Google OR-Tools, however, runs more than five times faster than it. We also see that Google OR-Tools outperforms CW, PSO, ACO and ALNS on all the instances. In addition, our method outperforms the four benchmark approaches on all the small scale instances as well.

\begin{table}[htbp]
  \centering
  \caption{Numerical results of the compared approaches on small scale instances. }
    \begin{tabular}{l|rr|rr|rr}
    \toprule
          & \multicolumn{2}{c|}{20 Nodes} & \multicolumn{2}{c|}{50 Nodes} & \multicolumn{2}{c}{100 Nodes} \\
    \midrule
          & \multicolumn{1}{l}{Cost/h} & \multicolumn{1}{l|}{Time/s} & \multicolumn{1}{l}{Cost/h} & \multicolumn{1}{l|}{Time/s} & \multicolumn{1}{l}{Cost/h} & \multicolumn{1}{l}{Time/s} \\
    \midrule
    OR-Tools & 2.378  & 0.054 & 3.072  & 0.374 & \textbf{3.900} & \textbf{1.71} \\
    \midrule
    Clarke-Wright & 2.428  & 0.001 & 3.169  & 0.009 & 4.069  & 0.055 \\
    PSO   & 2.386  & 16.3  & 3.286  & 40.66 & 4.506  & 82.4 \\
    ACO   & 2.456  & 2.86  & 3.397  & 11.41 & 4.589  & 35.49 \\
    ALNS  & 2.392  & 12.1  & 3.331  & 24.81 & 4.458  & 142 \\
    \midrule
    Our(Greedy) & 2.392  & 0.002 & 3.142  & 0.002 & 4.000  & 0.006 \\
    Our(Sample) & 2.375  & 0.039 & 3.058  & 0.116 & 3.928  & 0.304 \\
    Our(BS) & \textbf{2.367} & \textbf{0.013} & \textbf{3.050} & \textbf{0.056} & 3.936  & 0.209 \\
    \midrule
    Gap/speed* & best & 4.2   & best & 6.7   & 0.7\% & 5.6  \\
    \bottomrule
    \end{tabular}%
  \label{tab:small}%
\end{table}%

\begin{table}[htbp]
  \centering
  \caption{Numerical results of the compared approaches on large scale instances. }
    \begin{tabular}{l|rrr|rrr}
    \toprule
          & \multicolumn{3}{c|}{150 Nodes} & \multicolumn{3}{c}{200 Nodes} \\
    \midrule
          & \multicolumn{1}{l}{Cost/h} & \multicolumn{1}{l}{Time/s} & \multicolumn{1}{l|}{Solved} & \multicolumn{1}{l}{Cost/h} & Time/s & \multicolumn{1}{l}{Solved} \\
    \midrule
    OR-Tools & \textbf{4.492} & \textbf{341.6} & 69    & - & {1000} & 0 \\
    \midrule
    Clarke-Wright & 4.733  & 0.154 & 100   & 5.300  & {0.343} & 100 \\
    PSO   & 5.464  & 126.9 & 100   & 6.250  & {172.8} & 100 \\
    ACO   & 5.494  & 77.7  & 100   & 6.331  & {148.7} & 100 \\
    ALNS  & 5.506  & 295.8 & 100   & 5.728  & {1038} & 100 \\
    \midrule
    Our(Greedy) & 4.631  & 0.004 & 100   & 5.217  & {0.006} & 100 \\
    Our(Sample) & 4.542  & 0.577 & 100   & \textbf{5.097} & {\textbf{0.94}} & 100 \\
    Our(BS) & 4.564  & 0.479 & 100   & 5.142  & {0.253} & 100 \\
    \midrule
    Gap/speed* & 1.1\% & 592.0  &       & best & -     &  \\
    \bottomrule
    \end{tabular}%
  \label{tab:large}%
\end{table}%

Results of the compared approaches on 150- and 200-node instances are presented in \tref{tab:large}. The run-time of all solvers is limited within 1000 seconds. We also provide the number of instances that are solved within 1000 seconds. We can see that Google OR-Tools failed to solve all of the 150-node test instances within 1000 seconds. And it fails to solve all 200-node instances within this solving time limit. Even though our method performs 1.1\% worse than OR-Tools, it runs 592 times faster than OR-Tools on 150-node instances. Moreover, cost of the solutions found by our method is always lower than the other four compared algorithms. Clarke-Wright method, as the best-known heuristic for routing problem, can always produce a good solution in a reasonable run-time. But our method, by applying the greedy strategy, can outperform Clarke-Wright on all the small scale instances with identical run-time and all the large scale instances with much shorter run-time. On 200-node instances, our method with greedy strategy runs more than 50 times faster than all the compared algorithms while yielding the lowest cost. 

We observe from the numerical results on both small and large scale problem instances that:
\begin{itemize}
  \item In terms of optimality, Google OR-Tools and our method outperforms CW, PSO, ACO and ALNS on all the instances;
  \item In terms of optimality, Google OR-Tools outperforms our method on two (100- and 150-node instances) out of five instances. However, our method runs significantly faster than it on all of the instances; 
  \item Clarke-Wright shows a competitive performance to our method on 20-node instances in terms of run-time and optimality. However, our method performs obviously better on other instances; 
  \item The proposed method runs more than four times faster on small scale instances and more than 50 times faster on large scale instances than benchmark approaches while achieving the lowest cost on most instances.
\end{itemize} 

\begin{figure*}[htbp]
\centering
\subfigure[Greedy strategy]{\includegraphics[width=2.4in]{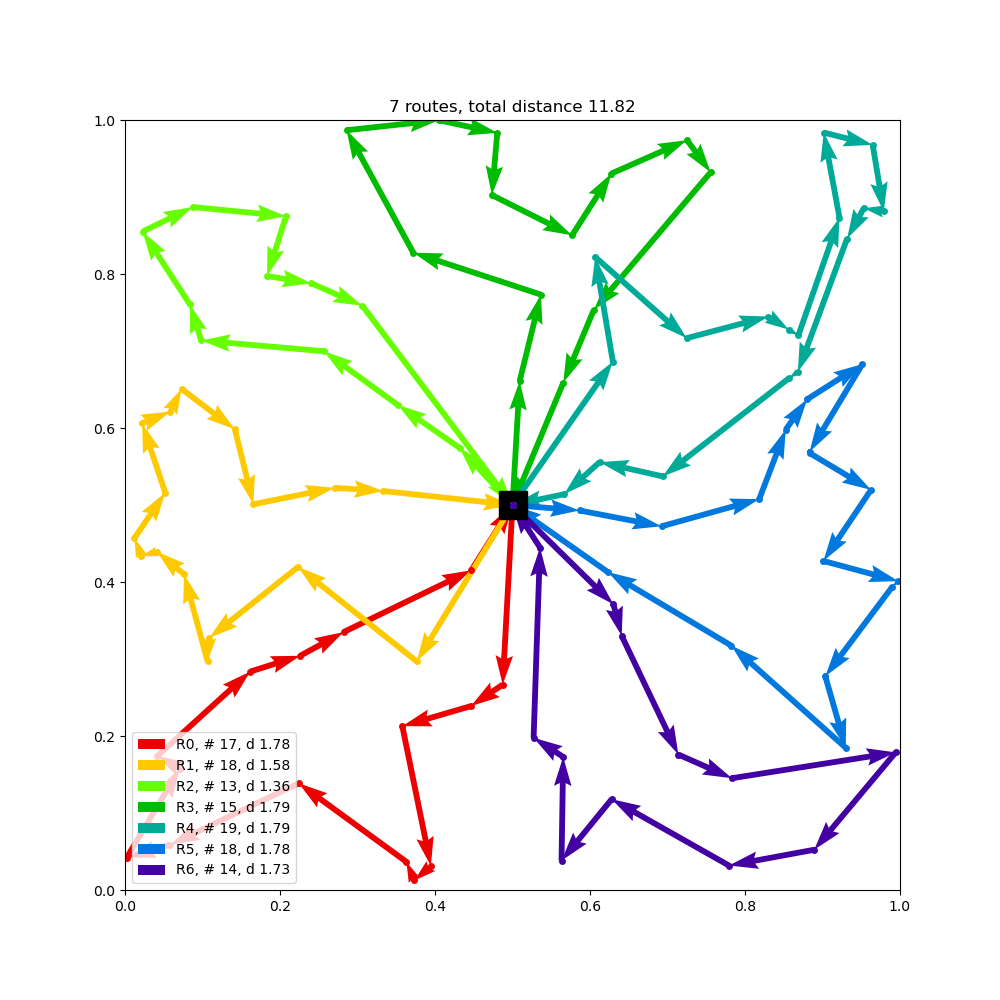}}
\hfil
\subfigure[Beam search strategy]{\includegraphics[width=2.4in]{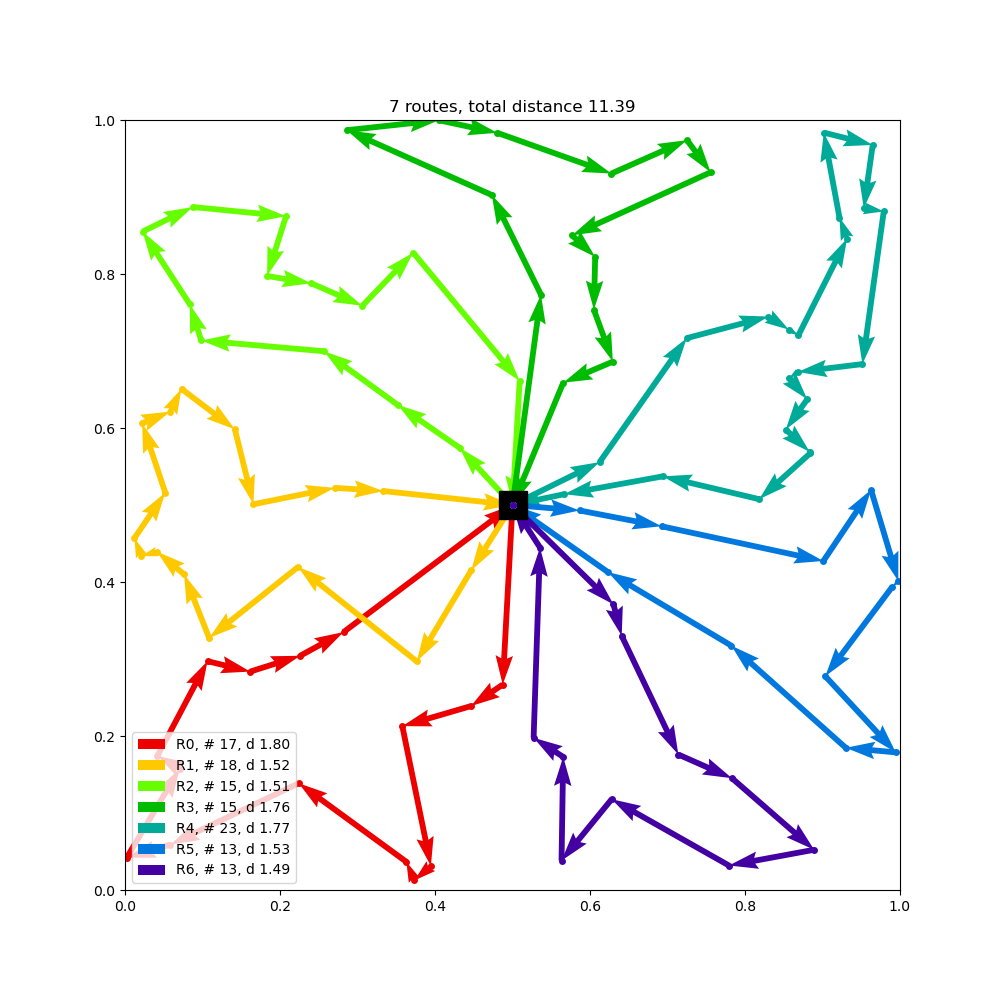}}

\caption{Solution found by the proposed method on a 100-node instance with greedy and beam search strategy.}
\label{fig:v2}
\end{figure*}

\begin{figure*}[htbp]
\centering
\subfigure[Sample strategy]{\includegraphics[width=2.4in]{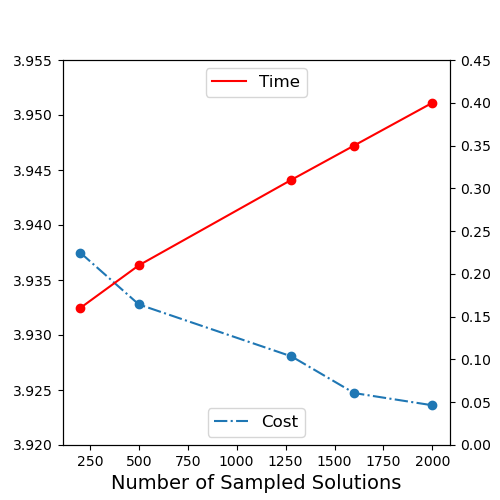}}
\hfil
\subfigure[Beam search strategy]{\includegraphics[width=2.4in]{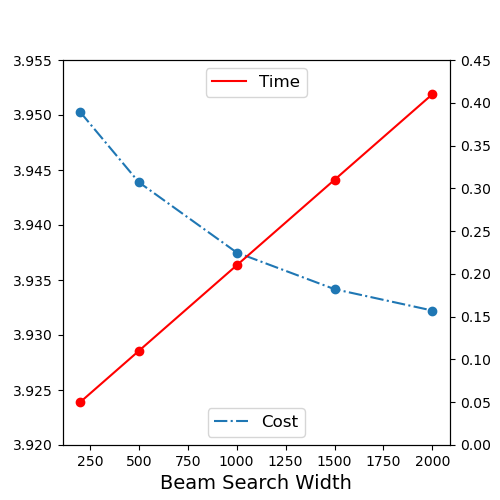}}

\caption{Performance analysis of sample and beam search strategies with different parameters.}
\label{fig:decode}
\end{figure*}

In addition, we analyze the performance of the decoding strategy. \fref{fig:v2} shows the solutions found by the proposed method on a 100-node instance with greedy and beam search strategy, respectively. We can see that the routes on the right of the figure are obviously improved by applying the beam search strategy. Meanwhile, the total run-time increases if using the sample and beam search strategy, since a set of solutions are generated. \fref{fig:decode} further analyzes the performance of sample and beam search strategies with different parameters. It is observed that the run-time of the sample and beam search strategy increases linearly with the increase of the number of sampled solutions and the beam search width, respectively. The cost of solutions found by applying the sample and beam search strategy is reduced correspondingly. Thus, we can change the parameters of the decoding strategies to achieve a trade-off between the solving speed and the solution quality, making the proposed method more practical for use.

\section{Conclusion}\label{con}
This paper proposes a new method to solve the UAV online routing problem with wireless power transfer. By using deep neural network to model this problem, the model can be trained offline and applied online, thus significantly reducing the online computational time. The model learns statistical similarities and data-driven heuristics from data while training. The learned knowledge is used to optimize the UAV routing problem online. This paradigm overcomes the limitations of traditional search-based approaches. Experiments strongly validates the effectiveness of the proposed approach, which runs significantly faster than the state-of-the-art Google OR-tools with identical solution quality. It also outperforms a variety of benchmark algorithms such as PSO and ACO. 

This study considers the power consumption of the UAV and the wireless charging process. Wireless power transfer can improve the endurance and range of the UAV in various applications when its battery capacity is limited. However, this study only considers the scenario where the UAV flies at a constant speed. It is worth investigating the acceleration and deceleration of the UAV when it arrives and leaves the base station. In addition, UAV routing problem with time window can be studied in the future using the deep learning method. 

Lastly, deep learning has shown very promising performance on a number of complex (combinatorial) optimization problems. However, the research in this filed is still in its infancy, we hope that this work can attract more and more researchers to investigate this topic.

%



\bibliographystyle{IEEEtran}
\bibliography{test}

\end{document}